\documentclass[10pt,twocolumn,letterpaper]{article}

\usepackage{cvpr}
\usepackage{times}
\usepackage{epsfig}
\usepackage{graphicx}
\usepackage{amsmath}
\usepackage{amssymb}
\usepackage{comment}
\usepackage{multirow}
\usepackage{xcolor}
\usepackage{caption} 
\usepackage[pagebackref=true,breaklinks=true,letterpaper=true,colorlinks,bookmarks=false]{hyperref}

\usepackage[accsupp]{axessibility}  

\cvprfinalcopy 

\definecolor{lightgray}{RGB}{220,220,220}
\definecolor{darkblue}{RGB}{0,0,127}
\definecolor{darkgreen}{RGB}{0,127,0}
\definecolor{darkred}{RGB}{200,0,0}

\ifcvprfinal\fi

\usepackage[affil-it]{authblk}

\usepackage{enumitem}

\title{The 7th AI City Challenge}

\begin{document}

\pagenumbering{gobble}

\author{
Milind Naphade$^1$ \hspace{0.9cm}
Shuo Wang$^1$ \hspace{0.9cm}
David C. Anastasiu$^2$ \hspace{0.9cm}
Zheng Tang$^1$ \\
Ming-Ching Chang$^3$ \hspace{0.9cm}
Yue Yao$^4$ \hspace{0.9cm}
Liang Zheng$^4$ \hspace{0.9cm}
Mohammed Shaiqur Rahman$^5$ \\
Meenakshi S. Arya$^5$ \hspace{0.9cm}
Anuj Sharma$^5$ \hspace{0.9cm}
Qi Feng$^7$ \hspace{0.9cm}
Vitaly Ablavsky$^8$ \\
Stan Sclaroff$^7$ \hspace{0.9cm}
Pranamesh Chakraborty$^6$ \hspace{0.9cm}
Sanjita Prajapati$^6$ \hspace{0.9cm}
Alice Li$^1$ \hspace{0.9cm} \\
Shangru Li$^1$ \hspace{0.9cm}
Krishna Kunadharaju$^1$ \hspace{0.9cm}
Shenxin Jiang$^1$ \hspace{0.9cm}
Rama Chellappa$^9$
} 
\affil{ 
$^1$ NVIDIA Corporation, CA, USA \hspace{0.8cm} 
$^2$ Santa Clara University, CA, USA \\ 
$^3$ University at Albany, SUNY, NY, USA \hspace{0.8cm} 
$^4$ Australian National University, Australia \\
$^5$ Iowa State University, IA, USA \hspace{0.8cm}
$^6$ Indian Institute of Technology Kanpur, India \\
$^7$ Boston University, MA, USA \hspace{0.8cm}
$^8$ University of Washington, WA, USA \\
$^9$ Johns Hopkins University, MD, USA
}

\maketitle

\begin{abstract}
The AI City Challenge's seventh edition emphasizes two domains at the intersection of computer vision and artificial intelligence - retail business and Intelligent Traffic Systems (ITS) - that have considerable untapped potential. The 2023 challenge had five tracks, which drew a record-breaking number of participation requests from 508 teams across 46 countries. Track 1 was a brand new track that focused on multi-target multi-camera (MTMC) people tracking, where teams trained and evaluated using both real and highly realistic synthetic data. Track 2 centered around natural-language-based vehicle track retrieval. Track 3 required teams to classify driver actions in naturalistic driving analysis. Track 4 aimed to develop an automated checkout system for retail stores using a single view camera. Track 5, another new addition, tasked teams with detecting violations of the helmet rule for motorcyclists. Two leader boards were released for submissions based on different methods: a public leader board for the contest where external private data wasn't allowed and a general leader board for all results submitted. The participating teams' top performances established strong baselines and even outperformed the state-of-the-art in the proposed challenge tracks.
\end{abstract}

\section{Introduction}

AI City is all about applying AI to improve the efficiency of operations in all physical environments. This manifests itself in reducing friction in retail and warehouse environments supporting speedier check-outs. It also manifests itself in improving transportation outcomes by making traffic more efficient and making roads safer. The common thread in all these diverse uses of AI is the extraction of actionable insights from a plethora of sensors through real-time streaming and batch analytics of the vast volume and flow of sensor data, such as those from cameras. The 7th edition of the AI City Challenge specifically focuses on problems in two domains where there is tremendous unlocked potential at the intersection of computer vision and artificial intelligence – retail business and Intelligent Traffic Systems (ITS). We solicited original contributions in these and related areas where computer vision, natural language processing, and deep learning have shown promise in achieving large-scale practical deployment that will help make our environments smarter and safer.

To accelerate the research and development of techniques, the 7th edition of this Challenge has pushed the research and development in multiple directions. We released a brand new dataset for multi-camera people tracking where a combination of real and synthetic data were provided for training and evaluation. The synthetic data were generated by the NVIDIA Omniverse Platform~\cite{nvidia_omniverse} that created highly realistic characters and environments as well as a variety of random lighting, perspectives, avatars, etc.  We also expanded the diversity of traffic related tasks such as helmet safety and the diversity of datasets including data from traffic cameras in India.

The five tracks of the AI City Challenge 2023 are summarized as follows:

\begin{itemize}[leftmargin=12pt] 

\item \textbf{Multi-target multi-camera (MTMC) people tracking:}
The teams participating in this challenge were provided with videos from various settings, including warehouse footage from a building and synthetically generated data from multiple indoor environments. The primary objective of the challenge is to track people as they move through the different cameras' fields of view. 

\item \textbf{Tracked-vehicle retrieval by natural language descriptions:} 
In this challenge track, the participating teams were asked to perform retrieval of tracked vehicles in the provided videos based on natural language (NL) descriptions. 184 held-out NL descriptions together with 184 tracked vehicles in 4 videos were used as the test set. 

\item  \textbf{Naturalistic driving action recognition:}
In this track, teams were required to classify 16 distracted behavior activities performed by the driver, such as texting, phone call, reaching back, {\em etc.} The synthetic distracted driving (\textit{SynDD2}) dataset~\cite{https://doi.org/10.48550/arxiv.2204.08096} used in this track was collected using three cameras located inside a stationary vehicle.

\item \textbf{Multi-class product recognition \& counting for automated retail checkout:} In this track, the participating teams were provided with synthetic training data only, to train a competent model to identify and count products when they move along a retail checkout lane. 

\item \textbf{Detecting violation of helmet rule for motorcyclists:}
Motorcycles are among the most popular modes of transportation, particularly in developing countries such as India. In many places, wearing helmets for motorcycle riders is mandatory as per traffic rules, and thus automatic detection of motorcyclists without helmets is one of the critical tasks to enforce strict regulatory traffic safety measures. The teams were requested to detect if the motorcycle riders were wearing a helmet or not.

\end{itemize}

Similar to previous AI City Challenges, there was considerable interest and participation in this year's event. Since the release of the challenge tracks in late January, we have received participation requests from 508 teams, representing a 100\% increase compared to the 254 teams that participated in 2022. The participating teams hailed from 46 countries and regions worldwide. In terms of the challenge tracks, there were 333, 247, 271, 216, and 267 teams participating in tracks 1 through 5, respectively. This year, 159 teams signed up for the evaluation system, up from 147 the previous year. Of the five challenge tracks, tracks 1, 2, 3, 4, and 5 received 44, 20, 42, 16, and 53 submissions, respectively.

The paper presents an overview of the 7th AI City Challenge's preparation and results. The following sections detail the challenge's setup ($\S$\ref{sec:challenge:setup}), the process for preparing challenge data ($\S$\ref{sec:dataset}), the evaluation methodology ($\S$\ref{sec:eval}), an analysis of the results submitted ($\S$\ref{sec:results}), and a concise discussion of the findings and future directions ($\S$\ref{sec:conclusion}).

\section{Challenge Setup}
\label{sec:challenge:setup}

\begin{figure*}[tbh]
  \includegraphics[width=1.0\linewidth]{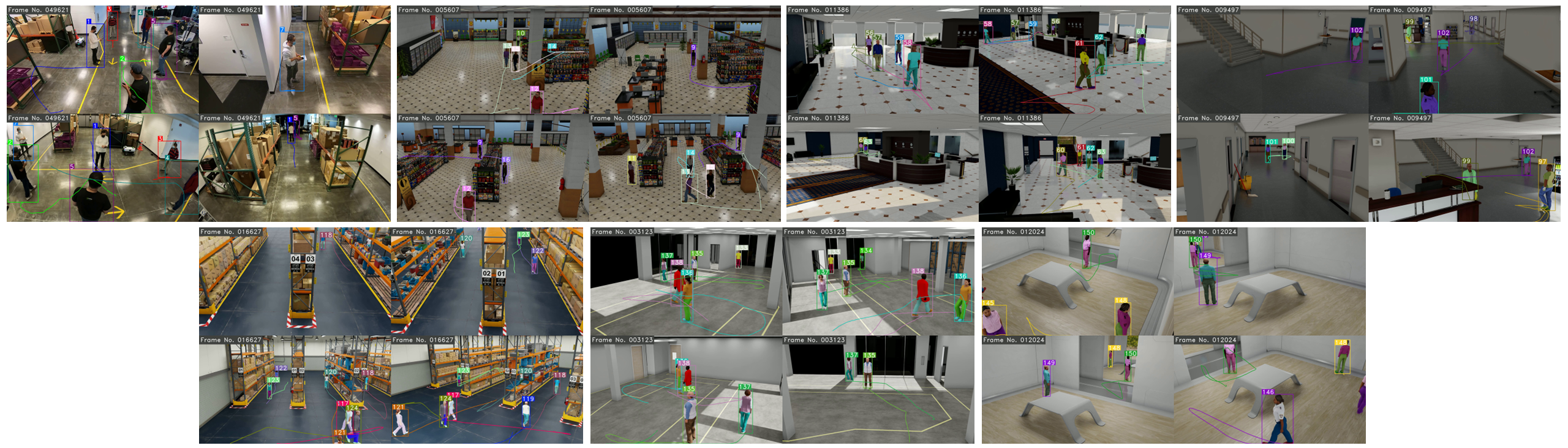}
  \caption{
    The MTMC people tracking dataset for Track 1 contains 22 subsets from 7 environments, including a real environment and six synthetic environments. The figure contains 4 sampled frames from each of the 7 environments.
  }
  \label{fig:mtmc-people-tracking}
  \vspace{-0.2 in}
\end{figure*}

The 7th AI City Challenge followed a format similar to previous years, with training and test sets released to participants on January 23, 2023, and all challenge track submissions due on March 25, 2023. Competitors vying for prizes were required to release their code for validation and make their code repositories publicly accessible, as we anticipated that the winners would make significant contributions to the community and the knowledge base. It was also necessary for the results on the leader boards to be reproducible without the use of external private data.

\textbf{Track 1: MTMC People Tracking.}
Teams were tasked with tracking people through multiple cameras by utilizing a blend of both real and synthetic data. This challenge differs significantly from previous years' vehicle multi-camera tracking, due to the unique features of an indoor setting, overlapping field of views, and the combination of real and synthetic data. The team that could achieve the most accurate tracking of people appearing in multiple cameras was declared the winner. In case of a tie, the winning algorithm was chosen to be the one that required the least amount of manual intervention.

\textbf{Track 2: Tracked-Vehicle Retrieval by Natural Language Descriptions.}
In this challenge track, teams were asked to perform tracked-vehicle retrieval given vehicles that were tracked in single-view videos and corresponding NL descriptions of the tracked vehicles. This track presents distinct and specific challenges which require the retrieval models to consider both relation contexts between vehicle tracks and motion within each track. Following the same evaluation setup used in previous years, the participating teams ranked all tracked vehicles for each NL description and the retrieval performance of the submitted models were evaluated using Mean Reciprocal Rank (MRR).

\textbf{Track 3: Naturalistic Driving Action Recognition.}
Based on about 30 hours of videos collected from 30 diverse drivers, each team was asked to submit one text file containing the details of one identified activity on each line. The details included the start and end times of the activity and corresponding video file information. Teams' performance was based on model activity identification performance, measured by the average activity overlap score, and the team with the highest average activity overlap score was declared the winner for this track.

\textbf{Track 4: Multi-Class Product Recognition \& Counting for Automated Retail Checkout.}
Participant teams were asked to report the object ID as well as the timestamp when a retail staff moved retail objects across the area of interest in pre-recorded videos. This track involves domain adaptation, as teams were required to perform domain transfer from synthetic data to real data to complete this challenge. Only synthetic data were provided for training. 

\textbf{Track 5: Detecting Violation of Helmet Rule for Motorcyclists.}
In this track, teams were requested to detect motorcycle drivers and passengers with or without helmet, based on the traffic camera video data obtained from an Indian city. Motorcycle drivers and passengers were treated as separate object entities in this track. The dataset included challenging real-world scenarios, such as poor visibility due to low-light or foggy conditions, congested traffic conditions at or near traffic intersections,  {\em etc.}

\section{Datasets}
\label{sec:dataset}

\begin{figure}[t]
\centering
\includegraphics[width=0.47\textwidth]{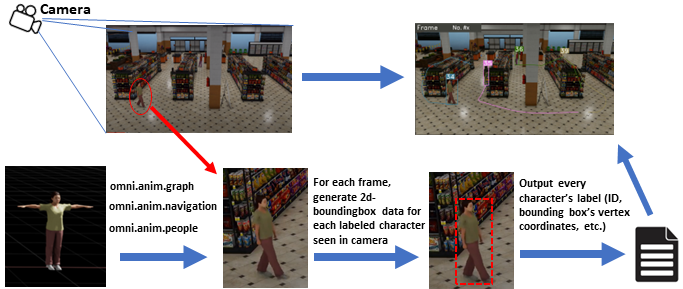}
\caption{The framework of MTMC synthetic data generation. The Omniverse Replicator collects each character's labels and converts them to a format usable by DNNs. The ``omni.anim.people'' extension is used to generate the scenario for each synthetic environment and enable each character to perform a sequence of movement.}
\label{fig:sdg-mtmct}
\end{figure}

The data for Track 1 were collected from multiple cameras in both real-world and synthetic settings. We used the NVIDIA Omniverse Platform~\cite{nvidia_omniverse} to create a large-scale synthetic animated people dataset, which was used for training and testing alongside the real-world data. For Track 2, we collected data from traffic cameras in several intersections of a mid-sized U.S. city and provided manually annotated NL descriptions. Track 3 participants were given synthetic naturalistic data of a driver from three camera locations inside the vehicle, where the driver was pretending to be driving. Track 4 involved identifying/classifying products held by a customer in front of a checkout counter, even when they were visually similar or occluded by hands and other objects. Synthetic images were provided for training, and evaluations were conducted on real test videos. Finally, Track 5 featured data from various locations in an Indian city, where we annotated each motorcycle with bounding box information and whether the riders were wearing helmets. In all cases, privacy was addressed by redacting vehicle license plates and human faces.

Specifically, we have provided the following datasets for the challenge this year: 
(1) The MTMC People Tracking dataset for Track 1, (2) \textit{CityFlowV2}~\cite{Tang19CityFlow, Naphade19AIC19, Naphade20AIC20, Naphade21AIC21, Naphade22AIC22} and \textit{CityFlow-NL}~\cite{feng2021cityflownl} for Track 2 on tracked-vehicle retrieval by NL descriptions, (3) \textit{SynDD2} for Track 3 on naturalistic driving action recognition, (4) The \textit{Automated Retail Checkout (ARC)} dataset for Track 4 on multi-class product counting and recognition, and (5) The Bike Helmet Violation Detection dataset for Track 5.

\begin{table}[t]
\caption{List of distracted driving activities in the {\it SynDD2} dataset.}
\centering
\begin{tabular}{|c|c|}
\hline
Sr. no. & Distracted Driver Behavior        \\ \hline\hline
0       & Normal forward driving            \\ \hline
1       & Drinking                          \\ \hline
2       & Phone call (right)                \\ \hline
3       & Phone call (left)                 \\ \hline
4       & Eating                            \\ \hline
5       & Texting (right)                   \\ \hline
6       & Texting (left)                    \\ \hline
7       & Reaching behind                   \\ \hline
8       & Adjusting control panel           \\ \hline
9       & Picking up from floor (driver)    \\ \hline
10      & Picking up from floor (passenger) \\ \hline
11      & Talking to passenger at the right \\ \hline
12      & Talking to passenger at backseat  \\ \hline
13      & Yawning                           \\ \hline
14      & Hand on head                      \\ \hline
15      & Singing and dancing with music    \\ \hline
\end{tabular}
\label{tab:driving:activities}
\end{table}


\subsection{The MTMC People Tracking Dataset}

The MTMC people tracking dataset is a comprehensive benchmark that includes seven different environments. The first environment is a real warehouse setting, while the remaining six environments are synthetic and were created using the NVIDIA Omniverse Platform (see Figure~\ref{fig:mtmc-people-tracking}). The dataset comprises a total of 22 subsets, of which 10 are designated for training, 5 for validation, and 7 for testing. The dataset includes a total of 129 cameras, 156 people and 8,674,590 bounding boxes. To our knowledge, it is the largest benchmark for MTMC people tracking in terms of the number of cameras and objects. Furthermore, the total length of all the videos in the dataset is 1,491 minutes, and all the videos are available in high definition (1080p) at 30 frames per second, which is another notable feature of this dataset. Additionally, all the videos have been synchronized, and the dataset provides a top-view floorplan of each environment that can be used for calibration. 

The ``Omniverse Replicator'' demonstrated in Figure~\ref{fig:sdg-mtmct} is a framework that we used to facilitate character labeling and synthetic data generation. It stores the rendered output of the camera, annotates it, and converts the data into a usable format for Deep Neural Networks (DNNs). To capture the spatial information of characters in each frame, Omniverse Replicator records 2D bounding boxes when characters are mostly visible in the camera frame, with their bodies or faces unoccluded. To simulate human behavior in synthetic environments, the ``omni.anim.people'' extension in Omniverse is used. This extension is specifically designed for simulating human activities in various environments such as retail stores, warehouses, and traffic intersections. It allows characters to perform predefined actions based on an input command file, which provides realistic and dynamic movements for each character in the scene.

\subsection{The {\bf \textit{CityFlowV2}} and {\bf \textit{CityFlow-NL}} Dataset}

The \textit{CityFlowV2} dataset comprises 3.58 hours (215.03 minutes) of videos obtained from 46 cameras placed across 16 intersections. The maximum distance between two cameras in the same scene is 4 km. The dataset covers a wide range of locations, including intersections, stretches of roadways, and highways, and is divided into six scenes, with three used for training, two for validation, and one for testing. The dataset contains 313,931 bounding boxes for 880 distinct annotated vehicle identities, with only vehicles passing through at least two cameras annotated. Each video has a resolution of at least 960p and is at 10 frames per second. Additionally, each scene includes an offset from the start time that can be used for synchronization. 

The \textit{CityFlow-NL} dataset~\cite{feng2021cityflownl} is annotated based on a subset of the \textit{CityFlowV2} dataset, comprising 666 target vehicles, with 3,598 single-view tracks from 46 calibrated cameras, and 6,784 unique NL descriptions. Each target vehicle is associated with at least three crowd-sourced NL descriptions, which reflect the real-world variations and ambiguities that can occur in practical settings. The NL descriptions provide information on the vehicle's color, maneuver, traffic scene, and relationship with other vehicles. We used the \textit{CityFlow-NL} benchmark for Track 2 in a single-view setup. Each single-view tracked vehicle is paired with a query consisting of three distinct NL descriptions for the training split. The objective during evaluation is to retrieve and rank tracked vehicles based on the given NL queries. This variation of the \textit{CityFlow-NL} benchmark includes 2,155 vehicle tracks, each associated with three unique NL descriptions, and 184 distinct vehicle tracks, each with a corresponding set of three NL descriptions, arranged for testing and evaluation.

\subsection{The {\bf \textit{SynDD2}} Dataset}

 \textit{SynDD2} \cite{https://doi.org/10.48550/arxiv.2204.08096} consists of 150 video clips in the training set and 30 videos in the test set. The videos were recorded at 30 frames per second at a resolution of $1920\times 1080$ and were manually synchronized for the three camera views~\cite{9857426}. Each video is approximately 9 minutes in length and contains all 16 distracted activities shown in Table~\ref{tab:driving:activities}. These enacted activities were executed by the driver with or without an appearance block such as a hat or sunglasses in random order for a random duration. There were six videos for each driver: three videos in sync with an appearance block and three other videos in sync without any appearance block.

\subsection{The {\bf \textit{Automated Retail Checkout (ARC)}} Dataset}

Inherited from the last year's challenge~\cite{Naphade22AIC22}, the {\it Automated Retail Checkout (ARC)} dataset includes two parts: synthetic data for model training and real-world data for model validation and testing.

\begin{figure}[t]
\centering
\includegraphics[width=0.47\textwidth]{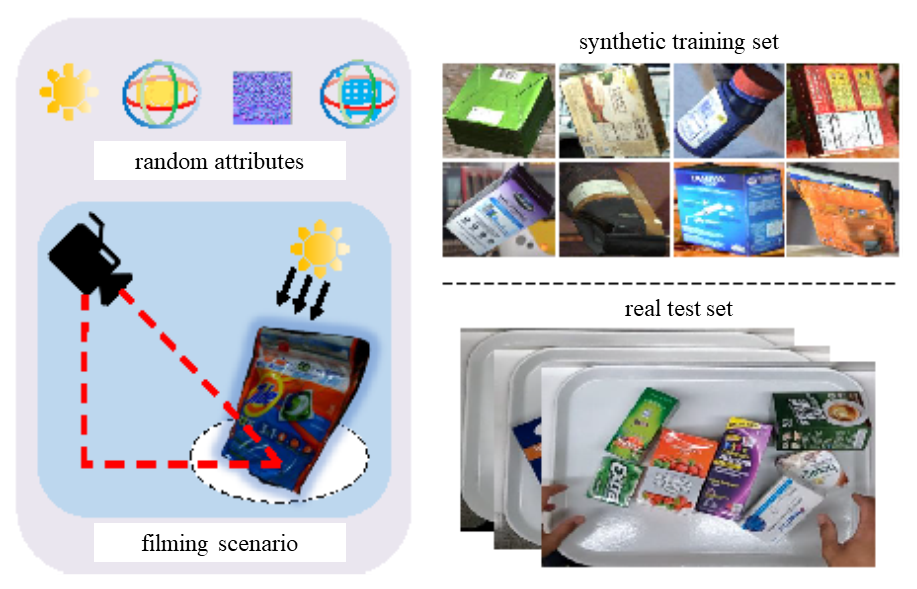}
\caption{The {\it Automated Retail Checkout (ARC)} dataset.}
\label{img:retail_datasets}
\end{figure}

The synthetic data was created using the pipeline from~\cite{yao2022attribute}. Specifically, we collected 116 scans of real-world retail objects obtained from supermarkets in 3D models. Object classes include daily necessities, food, toys, furniture, household, \etc. A total of $116,500$ synthetic images were generated from these $116$ 3D models. Images were filmed in a scenario demonstrated in Figure~\ref{img:retail_datasets}. Random attributes including random object placement, camera pose, lighting, and backgrounds were adopted to increase the dataset diversity. Background images were chosen from Microsoft COCO~\cite{lin2014microsoft}, which has diverse scenes suitable for serving as natural image backgrounds. This year we further provided the 3D models and Unity-Python interface for participating teams, so they could create more synthetic data if needed.

In our test scenario, the camera was mounted above the checkout counter and facing straight down, while a customer was enacting a checkout action by ``scanning'' objects in front of the counter in a natural manner. Several different customers participated, where each of them scanned slightly differently. There was a shopping tray placed under the camera to indicate where the AI model should focus. In summary, we obtained approximately $22$ minutes of videos, and the videos were further split into \textit{testA} and \textit{testB} sets such that \textit{testA} accounts for $40\%$ of the time and \textit{testB}, which accounts for $60\%$ was reserved for testing and determining the ranking of participating teams, accounts for the remainder. 

\subsection{The Bike Helmet Violation Detection Dataset}

The dataset was obtained from various locations of an Indian city. There are 100 videos in the training dataset and 100 videos in test dataset. Each video is 20 seconds in duration, recorded at 10 fps, at 1080p resolution. All pedestrian faces and vehicle license plates were redacted. There were 7 object classes annotated in the dataset, including motorbike, \textit{DHelmet} (Driver with helmet), \textit{DNoHelmet} (Driver without helmet), \textit{P1Helmet} (Passenger 1 with helmet), \textit{P1NoHelmet} (Passenger 1 without helmet), \textit{P2Helmet} (Passenger 2 with helmet), \textit{P2NoHelmet} (Passenger 2 without helmet). Bounding boxes were restricted to have a minimum height and width of 40 pixels, similar to the KITTI dataset~\cite{geiger2012we}. Further, an object was annotated if at least 40\% of the object was visible. The training dataset consists of a total of $\sim 65,000$ annotated objects.

\section{Evaluation Methodology}
\label{sec:eval}

As in previous AI City Challenges~\cite{Naphade18AIC18,Naphade19AIC19,Naphade20AIC20,Naphade21AIC21,Naphade22AIC22}, teams were encouraged to submit multiple solutions to our \textbf{online evaluation system} that automatically evaluated the performance of those solutions and provided that feedback both to the submitting team and other teams participating in the challenge. The top three results for each track were shown in an anonymized leaderboard as a way to incentivize teams to improve their methods. Team submissions were limited to 5 per day and a total of 20--40 submissions per track, depending on the track. Any erroneous submissions, i.e., those that contained a format or evaluation error, did not count against a team's daily or maximum submission totals. To discourage excessive fine-tuning of results or methods to improve performance, the results posted prior to the end of the challenge were computed on a 50\% random subset of the test set for each track, with the understanding that submitted methods should be generalizable and also perform well on the full test set. At the end of the competition, the full leader board with scores computed on the entire test set for each track was revealed.
Teams competing for the challenge prizes submitted results to the {\bf Public} leader board and were not allowed to use external private data or manual labeling on the test sets to fine-tune the performance of their models. Other teams were allowed to submit to a separate {\bf General} leader board.

\subsection{Track 1 Evaluation}

Similar to Track 3 in our 2021 Challenge~\cite{Naphade21AIC21} and Track 1 in our 2022 Challenge~\cite{Naphade22AIC22}, Track 1 was evaluated based on the IDF1 score~\cite{Ristani16Performance}, which measures the ratio of correctly identified detections over the average number of ground truths and computed detections. 

\subsection{Track 2 Evaluation} 
\label{sec:track2:eval}

Track 2 was originally inaugurated as Track 5 in our 2021 Challenge~\cite{Naphade21AIC21} and was reprised as Track 2 in our 2022 Challenge~\cite{Naphade22AIC22}. We used MRR as the effectiveness measure for this track, which is a standard metric for retrieval tasks~\cite{manning2008introduction}. In addition, the evaluation server provided teams with Recall@5, Recall@10, and Recall@25 results for their submissions, but these measures were not used in the ranking. 

\subsection{Track 3 Evaluation}
\label{sec:track1:eval}

While Track 3 is a reprisal of Track 3 in our 2022 Challenge~\cite{Naphade22AIC22}, we modified the evaluation measure to better account for activities that were correctly identified by teams during only a portion of the activity duration. Starting this year, Track 3 performance is based on model activity identification performance, measured by the average activity overlap score, which is defined as follows. Given a ground-truth activity $g$ with start time $gs$ and end time $ge$, let $p$ be its \textit{closest predicted activity} if it has the same class as $g$ and the highest overlap score $os$ among all activities that have overlap with $g$, with the added condition that its start time $ps$ and end time $pe$ are in the range $[gs – 10s, gs + 10s]$ and $[ge - 10s, ge + 10s]$, respectively. The overlap between $g$ and $p$ is defined as the ratio between the time intersection and the time union of the two activities, i.e.,
\[
os(p,g) = \frac{\max(\min(ge,pe)-\max(gs,ps), 0)}{\max(ge,pe)-\min(gs,ps)}.
\]
After matching each ground truth activity with at most one predicted activity and processing them in the order of their start times, all unmatched ground-truth activities and all unmatched predicted activities are assigned an overlap score of 0. The final score is the average overlap score among all matched and unmatched activities.

\subsection{Track 4 Evaluation}
\label{sec:track4:eval}

Evaluation for Track 4 was done using the same methodology as in Track 4 in our 2022 Challenge~\cite{Naphade22AIC22}, when this problem was first introduced in our Challenge. Performance was measured based on model identification performance, measured by the F1-score. To improve the resolution of the matches, the submission format was updated for this track to include frame IDs when the object was counted, rather than timestamps (in second), which previously led some teams to miss some predictitons in the last challenge due to reporting time in integers rather than floats.

\subsection{Track 5 Evaluation}
\label{sec:track5:eval}
Track 5 was evaluated based on mean Average Precision (mAP) across all frames in the test videos, as defined in the PASCAL VOC 2012 competition~\cite{pascal-voc-2012}. The mAP score computes the mean of average precision (the area under the Precision-Recall curve) across all the object classes. Bounding boxes with a height or width of less than 40 pixels and those that overlapped with any redacted regions in the frame were filtered out to avoid false penalization.

\section{Challenge Results}
\label{sec:results}

Tables~\ref{table:1}--\ref{table:5} summarize the leader boards for Tracks 1--5, respectively.

\subsection{Summary for the Track 1 Challenge}

\begin{table}[t]
\caption{Summary of the Track 1 leader board.}
\label{table:1}
\centering
\footnotesize
\begin{tabular}{|c|c|c|c|}
\hline
Rank & Team ID & Team & Score (IDF1) \\
\hline\hline
1 & 6 & UW-ETRI~\cite{UW23MTMCT} & {\bf 0.9536} \\
\hline
2 & 9 & HCMIU~\cite{HCMIU23MTMCT} & 0.9417 \\
\hline
3 & 41 & SJTU-Lenovo~\cite{SJTU23MTMCT} & 0.9331 \\
\hline
4 & 51 & Fraunhofer IOSB~\cite{Fraunhofer23MTMCT} & 0.9284 \\
\hline
5 & 113 & HUST~\cite{HUST23MTMCT} & 0.9207 \\
\hline
10 & 38 & Nota~\cite{Nota23MTMCT} & 0.8676 \\
\hline
13 & 20 & SKKU~\cite{SKKU23MTMCT} & 0.6171 \\
\hline
\end{tabular}
\vspace{-0.4cm}
\end{table}

Most teams followed the typical workflow of MTMC tracking, which consists of several components. (1) The first component is object detection, where all teams adopted YOLO-based models~\cite{bochkovskiy2020yolov5}. (2) Re-identification (ReID) models were used to extract robust appearance features. The top-performing team~\cite{UW23MTMCT} used OSNet~\cite{zhou2020omni}. The teams from HCMIU~\cite{HCMIU23MTMCT} and Nota~\cite{Nota23MTMCT} used a combination of multiple architectures and bag of tricks~\cite{zhou2020omni}. The HUST team~\cite{HUST23MTMCT} employed TransReID-SSL~\cite{wu2021self} pretrained on LUperson~\cite{zhong2019unsupervised}. (3) Single-camera tracking is critical for building reliable tracklets. Most teams used SORT-based methods, such as BoT-SORT~\cite{jiang2020bot} used in~\cite{UW23MTMCT} and~\cite{Nota23MTMCT}, and DeepSORT~\cite{Wojke17} used in~\cite{HCMIU23MTMCT}. The teams from SJTU-Lenovo~\cite{SJTU23MTMCT} and HUST~\cite{HUST23MTMCT} used ByteTrack~\cite{wang2021box} and applied tracklet-level refinement. (4) The most important component is clustering based on appearance and/or spatio-temporal information. The teams from UW-ETRI~\cite{UW23MTMCT} and Nota~\cite{Nota23MTMCT} used the Hungarian algorithm for clustering, where the former proposed an anchor-guided method for enhancing robustness. Most other teams~\cite{HCMIU23MTMCT, SJTU23MTMCT, Fraunhofer23MTMCT} adopted hierarchical clustering. The HUST team~\cite{HUST23MTMCT} applied k-means clustering, assuming the number of people was known, and refined the results using appearance, spatio-temporal, and face information. Most teams conducted clustering on appearance and spatio-temporal distances independently, but the SJTU-Lenovo team~\cite{SJTU23MTMCT} proposed to combine the distance matrices through adaptive weighting for clustering, which yielded satisfactory accuracy. Moreover, some teams~\cite{SJTU23MTMCT, HUST23MTMCT} found that conducting clustering within each camera before cross-camera association led to better performance. The SKKU team~\cite{SKKU23MTMCT} proposed a different method than all other teams, leveraging only spatio-temporal information for trajectory prediction using the social-implicit model, and achieving cross-camera association by spectral clustering. Therefore, their method achieved a good balance between accuracy and computation efficiency. To refine the homography-projected locations, they made use of pose estimation, which was also considered by other teams~\cite{UW23MTMCT, Nota23MTMCT}.

\subsection{Summary for the Track 2 Challenge}

\begin{table}[t]
  \caption{Summary of the Track 2 leader board.}
  \label{table:2}
  \centering
  \footnotesize
  \begin{tabular}{|c|c|c|c|}
    \hline
    Rank & Team ID & Team & Score (MRR) \\
    \hline\hline
    1 & 9 & HCMIU~\cite{HCMIU23NLRetrieval} & {\bf 0.8263} \\
    \hline
    2 & 28 & Lenovo~\cite{Lenovo23NLRetrieval} & 0.8179 \\
    \hline
    3 & 85 & HCMUS~\cite{HCMUS23NLRetrieval} & 0.4795 \\
    \hline
  \end{tabular}
  \vspace{-0.4cm}
\end{table}


In Track 2, the HCMIU team~\cite{HCMIU23NLRetrieval} introduced an improved retrieval model that used CLIP to combine text and image information, an enhanced Semi-Supervised Domain Adaptive training strategy, and a new multi-contextual pruning approach, achieving first place in the challenge. The MLVR model~\cite{Lenovo23NLRetrieval} comprised a text-video contrastive learning module, a CLIP-based domain adaptation technique, and a semi-centralized control optimization mechanism, achieving second place in the challenge. The HCMUS team~\cite{HCMUS23NLRetrieval} proposed an improved two-stream architectural framework that increased visual input features, considered multiple input vehicle images, and applied several post-processing techniques, achieving third place in the challenge.

\subsection{Summary for the Track 3 Challenge}

\begin{table}[t]
\caption{Summary of the Track 3 leader board.}
\label{table:3}
\centering
\footnotesize
\begin{tabular}{|c|c|c|c|}
\hline
Rank & Team ID & Team & Score (activity overlap score) \\
\hline\hline
1 & 209 & Meituan~\cite{Meituan23ActionRecognition} & {\bf 0.7416} \\
\hline
2 & 60 & JNU~\cite{JNU23ActionRecognition} & 0.7041 \\
\hline
3 & 49 & CTC~\cite{CTC23ActionRecognition} & 0.6723 \\
\hline
5 & 8 & Purdue~\cite{PurdueToyota23ActionRecognition,Purdue23ActionRecognition} & 0.5921 \\
\hline
7 & 83 & Viettel~\cite{Viettel23ActionRecognition} & 0.5881 \\
\hline
8 & 217 & NWU~\cite{NWU23ActionRecognition} & 0.5426 \\
\hline
16 & 14 & TUE~\cite{TUE23ActionRecognition} & 0.4849 \\
\hline
\end{tabular}
\vspace{-0.4cm}
\end{table}

The methodologies of the top performing teams in Track 3 of the Challenge were based on the basic idea of activity recognition, which involved (1) classification of various distracted activities, and (2) Temporal Action Localization (TAL), which determines the start and end time for each activity.  The best performing team, Meituan~\cite{Meituan23ActionRecognition}, utilized a self-supervised pretrained large model for clip-level video recognition. For TAL, a non-trivial clustering and removing post-processing algorithm was applied. Their best score was 0.7416. The runner-up, JNU~\cite{JNU23ActionRecognition} used an action probability calibration module for activity recognition, and designed a category-customized filtering mechanism for TAL. The third-place team, CTC~\cite{CTC23ActionRecognition} implemented a multi-attention transformer module which combined the local window attention and global attention. Purdue~\cite{PurdueToyota23ActionRecognition,Purdue23ActionRecognition} developed FedPC, a novel P2P FL approach which combined continual learning with a gossip protocol to propagate knowledge among
clients.

\subsection{Summary for the Track 4 Challenge}

\begin{table}[t]
\caption{Summary of the Track 4 leader board.}
\label{table:4}
\centering
\footnotesize
\begin{tabular}{|c|c|c|c|}
\hline
Rank & Team ID & Team & Score (F1) \\
\hline\hline
1 & 33 & SKKU~\cite{SKKU23AutomatedCheckout} & {\bf 0.9792} \\
\hline
2 & 21 & BUPT~\cite{BUPT22AutomatedCheckout} & 0.9787 \\
\hline
3 & 13 & UToronto~\cite{UToronto23AutomatedCheckout} & 0.8254 \\
\hline
4 & 1 & SCU~\cite{SCU23AutomatedCheckout} & 0.8177 \\
\hline
5 & 23 & Fujitsu~\cite{Fujitsu23AutomatedCheckout} & 0.7684 \\
\hline
6 & 200 & Centific~\cite{Centific23AutomatedCheckout} & 0.6571 \\
\hline
\end{tabular}
\end{table}

In Track 4, a task that involves synthetic and real data, we saw most teams performing optimization on both the training/testing data and recognition models. Specifically, for optimizing the training data, domain adaptation was performed to make the training data become visually similar to the real targets. For example, several teams used real-world background images to generate new training sets to train their detection and segmentation networks~\cite{SKKU23AutomatedCheckout,Fujitsu23AutomatedCheckout,Centific23AutomatedCheckout}. To improve the quality of the test data, teams performed deblurring~\cite{SKKU23AutomatedCheckout,UToronto23AutomatedCheckout}, hand removing~\cite{SKKU23AutomatedCheckout,UToronto23AutomatedCheckout,Fujitsu23AutomatedCheckout,SCU23AutomatedCheckout,BUPT23AutomatedCheckout}, and inpainting~\cite{SKKU23AutomatedCheckout,UToronto23AutomatedCheckout,SCU23AutomatedCheckout}. For optimizing the recognition models, teams followed the detection-tracking-counting (DTC) framework~\cite{SKKU23AutomatedCheckout,BUPT23AutomatedCheckout,Fujitsu23AutomatedCheckout,SCU23AutomatedCheckout}. In detection, YOLO-based models~\cite{bochkovskiy2020yolov5} were most commonly used~\cite{SKKU23AutomatedCheckout,UToronto23AutomatedCheckout,Centific23AutomatedCheckout,SCU23AutomatedCheckout}, followed by DetectoRS~\cite{BUPT23AutomatedCheckout,Fujitsu23AutomatedCheckout}. In tracking,  DeepSORT~\cite{wojke2017simple} and its improved version StrongSORT~\cite{du2023strongsort} were mostly used~\cite{SKKU23AutomatedCheckout,UToronto23AutomatedCheckout,Centific23AutomatedCheckout,SCU23AutomatedCheckout,BUPT23AutomatedCheckout}. Some teams further improved tracking by proposing new association algorithms. For example, \cite{Fujitsu23AutomatedCheckout} proposed CheckSORT, which achieved higher accuracy than DeepSORT~\cite{wojke2017simple} and StrongSORT~\cite{du2023strongsort}. Given the tracklets obtained from association, counting/post-processing was applied to get the timestamps when the object was in the area of interest.

\subsection{Summary for the Track 5 Challenge}

\begin{table}[t]
\caption{Summary of the Track 5 leader board.}
\label{table:5}
\centering
\footnotesize
\begin{tabular}{|c|c|c|c|}
\hline
Rank & Team ID & Team & Score (mAP) \\
\hline\hline
1 & 58 & CTC~\cite{CTC23HelmetDetection} & {\bf 0.8340} \\
\hline
2 & 33 & SKKU~\cite{SKKU23HelmetDetection} & 0.7754 \\
\hline
3 & 37 & VNPT~\cite{VNPT23HelmetDetection} & 0.6997 \\
\hline
4 & 18 & UTaipei~\cite{UTaipei23HelmetDetection} & 0.6422 \\
\hline
7 & 192 & NWU~\cite{NWU23HelmetDetection} & 0.5861 \\
\hline
8 & 55 & NYCU~\cite{NYCU23HelmetDetection} & 0.5569 \\
\hline
\end{tabular}
\end{table}

In Track 5, most teams followed the typical approach of object detection and multiple object tracking, which consists of several components. (1) The first component is object detection, and most teams used an ensemble model~\cite{casado2020ensemble} to improve the performance and generalization. The top performing team~\cite{CTC23HelmetDetection} used the Detection Transformers with Assignment (Deta) algorithm~\cite{deta} with a Swin-L~\cite{swin-l} backbone, whereas the second ranked team, SKKU~\cite{SKKU23HelmetDetection}, used YOLOv8~\cite{golroudbari2023recent}. The team from VNPT~\cite{VNPT23HelmetDetection}, who secured the third position in the track, used two separate models for Helmet Detection for Motorcyclists and Head Detection for detecting the heads of individual riders. (2) Object association or identification was used to correctly locate the driver/passenger. Most teams used the SORT algorithm~\cite{sort}. The top team~\cite{CTC23HelmetDetection} used Detectron2~\cite{wu2019detectron2} pretrained on the COCO dataset~\cite{lin2014microsoft} to obtain the detected motorcycles and people, and SORT~\cite{sort} to predict their trajectories and record their motion direction. The team from SKKU~\cite{SKKU23HelmetDetection} built an identifier model over YOLOv8~\cite{golroudbari2023recent}, to differentiate between motorbikes, drivers, and passengers and stored the resulting confidence scores. The team from VNPT~\cite{VNPT23HelmetDetection} used an algorithm to calculate the overlap areas and relative positions of the bounding boxes with respect to the motorbikes. (3) Finally, Category Refine modules were used to generate the results and correct any misclassified classes. All teams used diverse approaches for this module. The winning team~\cite{CTC23HelmetDetection} used SORT to associate the detected objects in different frames and get the trajectories of motorcycles and people. The team from SKKU~\cite{SKKU23HelmetDetection} relied on confidence values to filter the detections. The team from VNPT~\cite{VNPT23HelmetDetection} created an algorithm which identified the directions of the motorbikes and positions of the drivers and passengers, and corrected misclassified objects.

\section{Discussion and Conclusion}
\label{sec:conclusion}

The 7th AI City Challenge has continued to garner significant interest from the global research community, both in terms of the quantity and quality of participants. We would like to share a few noteworthy observations from the event.

For Track 1, we introduced a new benchmark for MTMC people tracking, which combines both real and synthetic data. The current state-of-the-art has achieved over 95\% accuracy on this extensive dataset. However, there are still challenges that need to be addressed before these methods can be deployed in general scenarios in the real world. Firstly, all participating teams adopted different approaches and configurations to handle the real and synthetic data in the test set separately. Nevertheless, one of the primary goals of this dataset is to encourage teams to explore MTMC solutions through domain adaptation. Secondly, due to a large number of camera views (129), many teams were unable to calibrate all of them and make use of the spatio-temporal information. We hope this dataset will encourage research into efficient camera calibration techniques with minimal human intervention. In the future, the camera matrices for synthetic data will be automatically generated. Thirdly, most of the leading teams assumed that the number of people is known in their clustering algorithm, which may not be valid in real-world scenarios. In future challenges, we also aim to discourage the use of faces in identity association and encourage participants to focus on balancing accuracy and computational efficiency.

For Track2, participating teams in the challenge used various approaches based on CLIP~\cite{radford2021learning} to extract motion and visual appearance features for the natural language guided tracked-vehicle retrieval task. Teams also implemented post-processing techniques based on the NL query's relation and motion keywords to further enhance retrieval results. The proposed models for Track 2 showed significant improvements in retrieval performance compared to the state-of-the-art from the 6th AI City Challenge, achieving an MRR of 82.63\% that represents a 30\% relative improvement. 

In Track 3, teams worked on the {\it SynDD2}~\cite{https://doi.org/10.48550/arxiv.2204.08096} benchmark and considered it as a Driver Activity Recognition problem with the aim to design an efficient detection method to identify a wide range of distracted activities. This challenge addressed two problems, classification of driver activity as well as temporal localization to identify the start and end time. To this end, teams have spent significant efforts in optimizing algorithms as well as implementing pipelines for performance improvement. They tackled the problem by adopting techniques including vision transformers~\cite{JNU23ActionRecognition,tong2022videomae,Sun_2019_CVPR,Li_2022_CVPR} and action classifiers~\cite{9171561,10.1007/978-3-031-19772-7_29,Liang_2022_CVPR,Lin_2019_ICCV}. Both activity recognition and temporal action localization are still open research problems that require more in-depth studies. More clean data and ground-truth labels can clearly improve the development and evaluation of the research progress. We plan to increase the size and quality of the {\it SynDD2} dataset, with the hope that it will significantly boost future research in this regard. 

In Track 4, teams worked on retail object recognition and counting methods. Substantial efforts were made on optimizing both the data and recognition algorithms. This year, we observed that more teams began to prioritize data-centric improvements rather than solely focusing on model-centric improvements, which was a positive trend. Specifically, optimization was carried out on both synthetic training data and real testing data, with the aim of reducing the domain gap between them. In the training with synthetic data, we saw methods that made synthetic data more realistic, while in testing with real data, we saw methods for denoising and data cleaning. We also noted significant improvements in recognition algorithms, including the usage of the latest detection and association models. The leading team used both data-centric and model-centric methods, resulting in over 97\% accuracy on the \textit{testA} (validation set). Moving forward, we hope to see further studies on how data-centric methods can be utilized, as well as how they can collaborate with model-centric methods to achieve even higher accuracy.

In Track 5, teams were provided with a diverse and challenging dataset for detecting motorbike helmet violation from an Indian city. The current state-of-the-art model achieved 0.83 mAP~\cite{CTC23HelmetDetection} on this extensive dataset. The top teams tackled the problem by adopting state-of-the-art object detection along with ensemble techniques~\cite{casado2020ensemble} and object tracking to improve model accuracy.  

\section{Acknowledgment}

The datasets of the 7th AI City Challenge would not have been possible without significant contributions from the Iowa DOT and an urban traffic agency in the United States. This Challenge was also made possible by significant data curation help from the NVIDIA Corporation and academic partners at the Iowa State University, Boston University, Australian National University, and Indian Institute of Technology Kanpur. We would like to specially thank Paul Hendricks and Arman Toorians from the NVIDIA Corporation for their help with the retail dataset.


{\small
\bibliographystyle{ieee_fullname}
\bibliography{aicity23, aicity22, aicity21, aicity20, aicity19, aicity18, aicity17}
}

\end{document}